\documentclass[letterpaper, 10 pt, conference]{ieeeconf}  

\IEEEoverridecommandlockouts                              

\usepackage{graphics} 
\usepackage{epsfig} 
\usepackage{mathptmx} 
\usepackage{times} 
\usepackage{amsmath} 
\usepackage{amssymb}  

\usepackage{algorithm}
\usepackage{algorithmic}

\usepackage{hyperref}

\title{\LARGE \bf
Gaze-Aware Task Progression Detection Framework for Human-Robot Interaction Using RGB Cameras
}

\author{Linlin Cheng$^{1}$, Koen Hindriks$^{1}$, and Artem V. Belopolsky$^{2}$%
\thanks{} 
\thanks{$^{1}$First Author and Second Author are with Department of Computer Science, Vrije Universiteit Amsterdam, Amsterdam, The Netherlands.
        {\tt\footnotesize linlincheng.research@gmail.com;k.v.hindriks@vu.nl}}%
\thanks{$^{2} $Third Author is with Department of Movement Sciences, Vrije Universiteit Amsterdam, Amsterdam, The Netherlands.
        {\tt\footnotesize a.belopolskiy@vu.nl}}%
\thanks{This article has been accepted for publication in \textit{IEEE Robotics and Automation Letters}.
DOI: \url{https://doi.org/10.1109/lra.2026.3673990} © IEEE}
}

\begin{document}

\maketitle
\thispagestyle{empty}
\pagestyle{empty}

\begin{abstract}
In human–robot interaction (HRI), detecting a human’s gaze helps robots interpret user attention and intent. However, most gaze detection approaches rely on specialized eye-tracking hardware, limiting deployment in everyday settings. 
Appearance-based gaze estimation methods remove this dependency by using standard RGB cameras, but their practicality in HRI remains underexplored. We present a calibration-free framework detecting task progression when information is conveyed via integrated display interfaces. The framework uses only the robot’s built-in monocular RGB camera (640 × 480 resolution) and state-of-the-art gaze estimation to monitor attention patterns. It leverages natural behavior, where users shift focus from task interfaces to the robot’s face to signal task completion, formalized through three Areas of Interest (AOI): tablet, robot face, and elsewhere. Systematic parameter optimization identifies configurations balancing detection accuracy and interaction latency. We validate our framework in a “First Day at Work” scenario, comparing it to button-based interaction. Results show a task completion detection accuracy of 77.6\%. Compared to button-based interaction, the proposed system exhibits slightly higher response latency but preserves information retention and significantly improves comfort, social presence, and perceived naturalness.  Notably, most participants reported that they did not consciously use eye movements to guide the interaction, underscoring the intuitive role of gaze as a communicative cue.  This work demonstrates the feasibility of intuitive, low-cost, RGB-only  gaze-based HRI for natural and engaging interactions.

\vspace{1em} 

\end{abstract}



\section{INTRODUCTION}

Gaze is fundamental to human communication, conveying attention, intention, and social readiness \cite{rakovic2022gaze,rickert2019applications,admoni2017social}. However, its integration into HRI remains limited due to reliance on specialized hardware \cite{ban2023persistent, wang2023gaze}, which, despite decreasing costs, still hinders widespread adoption and alters natural user behavior through physical encumbrance. Alternative strategies also present notable limitations. Approaches approximating gaze through head orientation lack precision to capture subtle eye movements \cite{ivaldi2014robot}. Model-based techniques often rely on constrained head positions and high-quality imaging conditions \cite{ palinko2015eye}. In contrast, appearance-based methods offer a more flexible solution by leveraging standard RGB cameras to infer gaze direction from face images, removing dependency on specialized hardware and calibration procedures. However, their initially limited gaze range restricted HRI applications. Recent large-scale datasets have extended this range substantially, enabling development of state-of-the-art gaze estimation models with demonstrated potential for HRI  \cite{cheng2023boundary, cheng2024automating}.

Despite these advances, their sufficiency for enabling robust, real-time gaze-based interaction in practical HRI settings remains largely unexplored. Prior work has focused primarily on developing and benchmarking gaze estimation models \cite{biswas2021appearance, cheng2024appearance}, with limited attention to their integration into complete, ecologically valid HRI systems. Furthermore, many HRI studies depend on intrusive hardware like eye-tracking glasses ~\cite{shi2021gazeemd,yuan2019human,li20173}, which require calibration and explicitly inform users about detection mechanisms. These factors heighten awareness, disrupt natural interaction flow, and ultimately reduce ecological validity.

To address these challenges, we propose a low-cost, calibration-free framework designed for HRI scenarios where robots need to detect when users have completed visual tasks presented on integrated display devices. Our approach targets interaction paradigms in which robots present sequences of visual information (e.g., text, images, maps) and must determine the appropriate timing to progress to subsequent items. Instead of requiring users to physically approach and press a button, it leverages natural gaze behavior to trigger automatic progression.

The framework assumes that the robot incorporates a display device (such as a tablet) positioned spatially separate from its face, allowing the system to clearly distinguish between task-focused attention and social attention directed toward the robot. Our key insight builds on a simple but powerful observation: when completing visual tasks, users naturally shift their gaze from the task interface (the robot’s tablet) to the interactor’s eye region (the robot’s face) to signal task completion and readiness to hand over the turn.
This mirrors fundamental human communication patterns, where eye contact facilitates turn-taking and shared attention \cite{strongman1968dominance,degutyte2021role}.

The proposed framework defines three key AOIs: (1) the task-related area (tablet), (2) the interactor’s eye area (robot’s face), and (3) elsewhere (all other regions). We establish system parameters to differentiate genuine task engagement, task disengagement, and incidental gaze movements with a simple parameter configuration that balances detection reliability with temporal responsiveness through five optimized values.
To evaluate this approach, we employ the “First Day at Work” paradigm, a naturalistic, socially meaningful scenario, using the Pepper robot platform. We compare gaze-based progression against traditional button-based methods to systematically assess the feasibility of proposed framework in practical HRI applications.

This research makes several contributions:
\begin{enumerate}
    \item \textbf{Framework introduction:} We present a low-cost, calibration-free framework specifically designed for HRI scenarios where robots need to detect when users complete visual tasks on integrated display devices, using only the robot’s built-in low-resolution RGB camera.
    \item \textbf{Design Guidelines:} We establish simple yet effective parameter settings and design principles that achieve optimal balance between detection reliability and temporal responsiveness for real-time gaze-based interaction systems.
    \item \textbf{Empirical validation:} We provide a comprehensive evaluation of the proposed framework within the ``First Day at Work'' paradigm using the Pepper robot platform \cite{pandey2018mass}, demonstrating the feasibility and effectiveness of appearance-based gaze estimation in real-world HRI.
\end{enumerate}

\section{Related work}

Gaze estimation plays a critical role in HRI \cite{rakovic2022gaze, admoni2017social}. However, many existing gaze estimation approaches in HRI rely heavily on specialized hardware \cite{duarte2018action, wang2023gaze, ban2023persistent}. Which have been used to study visual attention, such as identifying which robot parts or interfaces attract users’ attention \cite{schreiter2023advantages} or to infer user intent in industrial and teleoperation tasks \cite{ghosh2023automatic}, as well as to analyze engagement and joint attention in social HRI scenarios.
Such hardware-based methods typically employ infrared illumination to track eye movements, sometimes combined with additional visible-light sources to capture corneal reflections. While these devices can achieve high accuracy, they require users to wear additional equipment and demand explicit calibration, introducing physical encumbrance and self-awareness, which is impractical in real world and can negatively impact natural social interaction ~\cite{risko2011eyes,wong2019eye}.

To avoid the use of additional hardware, many researchers have adopted head orientation as a proxy of gaze estimation, commonly referred to as “head gaze”. For instance, Ivaldi et al. \cite{ivaldi2014robot} used the iCub robot’s onboard camera to estimate a human partner’s attention based on head direction. However, head orientation alone is an unreliable proxy, as people can shift their gaze without moving their heads during social interaction \cite{palinko2016robot}. Model-based gaze estimation approaches attempt to overcome this limitation by using either 2D facial features (e.g., eye and facial landmarks) or 3D geometric models of the eye to infer gaze direction \cite{palinko2015eye}. Despite their effectiveness under constrained conditions, these methods are typically limited to a narrow range of head poses and often require frequent re-calibration, restricting their applicability in dynamic HRI scenarios.

In this context, appearance-based gaze estimation methods have emerged as a promising alternative. These approaches infer gaze direction directly from RGB images, eliminating the need for specialized hardware and explicit calibration procedures. However, their performance is strongly influenced by the characteristics of the datasets used for training. Over the years, some gaze datasets have been introduced, but many are designed for screen-based applications (e.g., desktops or smartphones) \cite{16huang2017tabletgaze16,17krafka2016eye17}, focusing on 2D gaze estimation, which is not suitable for natural HRI settings. Other datasets include only limited variations in head pose and gaze direction \cite{18zhang2017mpiigaze18,19fischer2018rt19,20funes2014eyediap20}, typically restricting yaw angles to within ±40°, whereas realistic social interactions may require yaw variations of at least ±80° relative to the camera. The recent release of large-scale datasets such as Gaze360 and ETH-XGaze \cite{kellnhofer2019gaze360, zhang2020eth} has significantly advanced appearance-based gaze estimation. These datasets cover wide head-pose variations, with yaw angles reaching ±120°. Models trained on Gaze360 have proven particularly suitable for HRI applications \cite{cheng2024automating, cheng2023boundary}, as the dataset was collected in both indoor and outdoor environments and spans interaction distances ranging from 1 m to 3 m—corresponding closely to typical social distances in HRI scenarios. Consequently, this work employs a 3D gaze estimation model trained on the Gaze360 dataset, enabling robust detection of gaze transitions between arbitrary task regions and generalization to varying human–robot distances. Prior research on appearance-based gaze estimation has largely focused on developing and benchmarking gaze estimation algorithms \cite{biswas2021appearance, cheng2024appearance, zhang2022gazeonce, balim2023efe}, comparatively little attention has been paid to their integration into complete, ecologically valid HRI systems. In this work, we therefore focus on exploring how to engineer around state-of-the-art gaze estimation algorithms and investigate their practical application within real-world HRI contexts.

\section{Method}

\begin{figure*}
  \centering
  \includegraphics[width=1\textwidth]{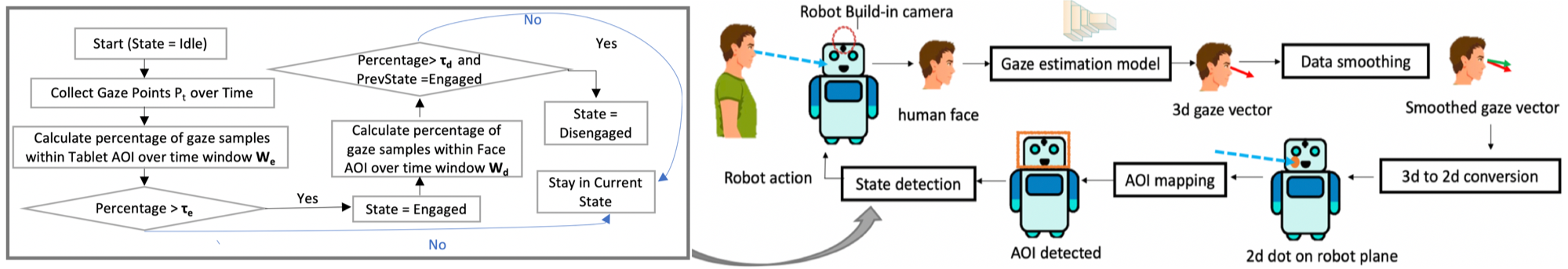}
  \caption{The overview of the proposed framework. The system uses the humanoid robot's built-in monocular camera to continuously track human facial features and estimate gaze direction using a pretrained deep learning model. The resulting 3D gaze vector undergoes smoothing filtration to reduce noise, then is mathematically projected onto the robot's 2D interaction plane and mapped to predefined AOIs. In the state detection algorithm(left flowchart), the user’s level of engagement was determined by analyzing their gaze patterns over time within predefined AOIs using a simple threshold-based approach. 
}
  \label{fig:method}
\end{figure*}

Figure \ref{fig:method} illustrates the presented framework: the robot’s camera captures the user’s gaze with a pretrained model, projects it onto the robot plane, and maps it to AOIs. A state detection algorithm then infers task readiness, enabling the robot to adapt its behavior. Interaction begins in an idle state: the user is considered engaged when gaze toward the tablet exceeds a threshold within a time window, and disengaged when gaze shifts to the robot’s face above a threshold after engagement. If conditions are not met, the system maintains the current state.
The code is publicly available at
\footnote{\label{github} 
\url{https://github.com/LinReseach/Gaze-Aware-Task-Framework}}.

\subsection{Gaze estimation model}


We employed the L2CS-Net model \cite{abdelrahman2023l2cs} for robust gaze estimation. L2CS-Net is a state-of-the-art convolutional neural network that predicts 3D gaze directions (yaw and pitch) from single RGB images in unconstrained real-world conditions. It uses a ResNet-50 backbone for feature extraction and dual fully-connected heads for yaw and pitch, improving accuracy by treating each dimension independently. The model achieves a mean angular error of 10.41° on the Gaze360 dataset \cite{kellnhofer2019gaze360} and is robust to variations in lighting, head pose, and facial appearance, making it suitable for human-robot interaction. Its efficiency enables real-time deployment, processing 640×480 RGB frames captured by the robot’s head camera in 126 ms on an RTX 3070 GPU, providing sufficient resolution for accurate facial feature detection while maintaining real-time performance.

\subsection{Data smoothing}
To mitigate noise and ensure stable gaze predictions, we applied a simple moving average filter with a window size of $N$ frames to the estimated gaze angles, which reduces frame-by-frame variability and enhances the reliability of task completion detection. 




\subsection{3D-to-2D Gaze Projection}

\begin{figure}[h]
  \centering
  \includegraphics[width=0.8\linewidth]{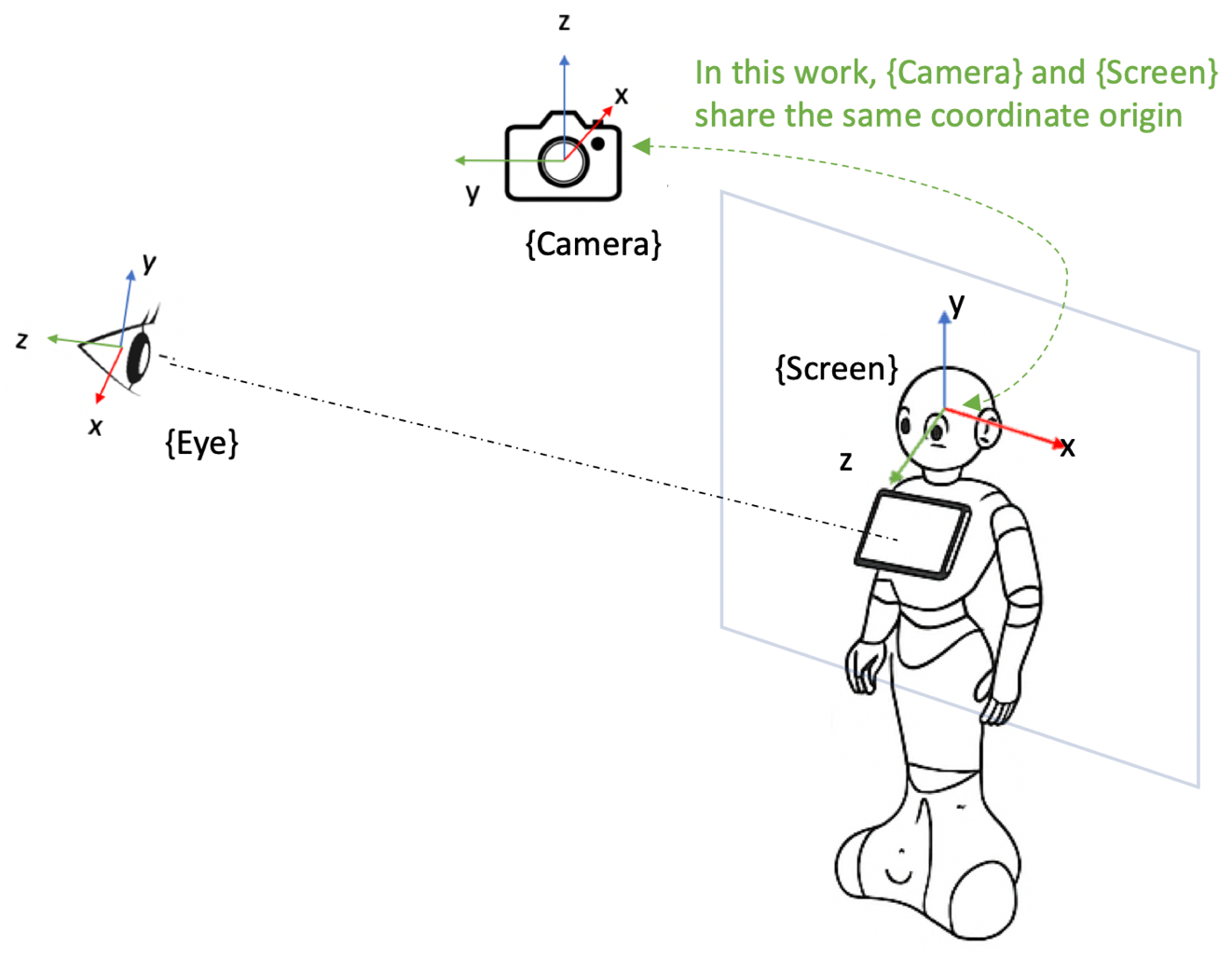}
  \caption{The relationship between 3D gaze directions and their projection
onto a 2D screen}
  \label{fig:3dto2dsetup}
\end{figure}
To project estimated yaw ($\theta$) and pitch ($\phi$) angles onto 2D screen coordinates, we establish a geometric transformation using screen, eye, and camera coordinate systems (Figure \ref{fig:3dto2dsetup}). The gaze direction vector in eye coordinates is:
\begin{equation}
\mathbf{v}_E = [\cos(\phi) \sin(\theta), -\sin(\phi), \cos(\phi) \cos(\theta)]
\end{equation}

The transformation involves four steps:

\textbf{Step 1:} Obtain screen pose $\{R_s, T_s\}$ and camera pose $\{R_c, T_c\}$ through calibration.

\textbf{Step 2:} Define screen plane with normal $n = R_s[:, 2]$ and point $T_s$:
\begin{equation}
n_x x + n_y y + n_z z = n \cdot T_s
\end{equation}

\textbf{Step 3:} Represent line of sight as:
\begin{equation}
\frac{x - x_o}{g_x} = \frac{y - y_o}{g_y} = \frac{z - z_o}{g_z}
\end{equation}
where $\mathbf{g} = R_c^{-1} \mathbf{v}_E$ is gaze direction and $\mathbf{o} = -R_c^{-1} T_c$ is gaze origin in camera coordinates.

\textbf{Step 4:} Compute intersection with screen plane and project to 2D coordinates:
\begin{equation}
p = (x, y), \quad \text{where} \quad (x, y, z) = R_s^{-1}(t - T_s) \quad \text{and} \quad z = 0
\end{equation}

\textbf{Implementation Configuration}
In our experimental setup, we utilize the Pepper robot's built-in head cameras for gaze tracking. To simplify the transformation process, the camera origin is aligned with the screen origin, resulting in the translation vector $T_s = [0, 0, 0]$.  The eye position is fixed throughout the task and kept the same for all participants, with a seat height of 130cm, chosen to approximate the eye level of an average seated adult.

\subsection{Area of Interest (AOI) Mapping}



After obtaining 2D gaze coordinates, each point is mapped to one of three predefined AOIs : the \textbf{Tablet AOI} (task-related screen region), the \textbf{Face AOI} (robot’s face region), and \textbf{Elsewhere}. The mapping function
$M: \mathbb{R}^2 \rightarrow \{\text{Tablet}, \text{Face}, \text{Elsewhere}\}$
assigns each gaze point $p = (x, y)$ as:

\begin{equation}
M(p) =
\begin{cases}
\text{Tablet} & \text{if } p \in \text{AOI}_{\text{tablet}} \\
\text{Face} & \text{if } p \in \text{AOI}_{\text{face}} \\
\text{Elsewhere} & \text{otherwise}
\end{cases}
\end{equation}

\subsection{State Detection Algorithm}  
The state detection algorithm infers the user’s engagement status from temporal gaze distributions over predefined Areas of Interest (AOIs). A threshold-based decision mechanism with sliding time windows is employed to ensure temporal stability.

\textbf{State Definitions}  

User states are defined based on the percentage of gaze samples within specific AOIs:

\begin{itemize}
    \item \textbf{Idle State (Initial)}: At the beginning of the interaction, the participant’s state is initialized as \textit{Idle}:
    \begin{equation}
        \text{State}_0 = \text{Idle}
    \end{equation}

    \item \textbf{Engaged State}: The user is classified as \textit{engaged} when the percentage of gaze samples directed toward the Tablet AOI exceeds a threshold $\tau_e$ over a time window $W_e$:
    \begin{equation}
        \text{State} = \text{Engaged} \iff 
        \frac{\sum_{t \in W_e} \mathbb{I}[M(P_t) = \text{Tablet}]}{|W_e|} > \tau_e
    \end{equation}

    \item \textbf{Disengaged State}: The user is classified as \textit{disengaged} when the percentage of gaze samples directed toward the Face AOI exceeds a threshold $\tau_d$ over a window $W_d$, provided that the previous state was \textit{Engaged}:
    \begin{equation}
    \text{State} = \text{Disengaged} \iff \frac{\sum_{t \in W_d} \mathbb{I}[M(P_t) = \text{Face}]}{|W_d|} > \tau_d
    \end{equation}
    \begin{equation}
    \land \text{PrevState} = \text{Engaged}
    \end{equation}
\end{itemize}

Here, $\mathbb{I}[\cdot]$ denotes the indicator function, $P_t$ is the gaze point at time $t$, and $M(\cdot)$ represents the AOI mapping function. An overview of the state detection process is shown in Figure~\ref{fig:method}.




    
    
    


\subsection{Parameter Configuration}
This framework relies on five key parameters: the Smoothing Window size ($N$), Engagement Confirmation Window ($W_e$), Disengagement Confirmation Window ($W_d$), Engagement Threshold ($\tau_e$), and Disengagement Threshold ($\tau_d$). To determine appropriate values for these parameters, we conducted a series of pretests. We first evaluated the accuracy of the pretrained gaze model by asking five participants to fixate on the tablet AOI and the robot’s face AOI for 5 seconds each, while the robot’s head camera recorded their faces. The model achieved a true positive rate of approximately 50\% for face AOI detection and 40\% for tablet AOI detection, leading us to set the engagement threshold $\tau_e = 40\%$ and the disengagement threshold $\tau_d = 50\%$. The remaining parameters---$N$, $W_e$, and  $W_d$---were selected jointly, as they collectively influence the framework's timing accuracy. We tested $N \in \{1,3,5,10,15\}$, $W_e \in \{0.5s, 1s, 1.5s, 2s, 3s\}$, and $W_d \in \{0.5s, 1s, 1.5s, 2s, 3s\}$ by having one participant complete six trials per configuration. 
The lower bound of 0.5s was chosen because our system, operating at the given frame rate, requires about 200~ms to process a single frame, making shorter windows impractical. Using 200ms as a window would rely on only one frame and be highly susceptible to noise and misclassification. A minimum of 0.5s ($\approx$ 2--3 frames) provides a more stable basis for detection while still keeping the system responsive. The upper bound of 3s was set to avoid excessive delays, as longer windows would result in noticeably late page turns and risk participant frustration. Timing-aware accuracy, defined as the percentage of trials where the system correctly detected task completion within a subjectively acceptable window (while minimizing both early and late detection errors), was used as the evaluation metric. The configuration $N = 3$, $W_e = 1s$, and $W_d = 1s$ achieved the highest timing-aware accuracy at 83\%. The final choice of parameters are shown in Table~\ref{tab:parameters}.

\begin{table}[h]
\centering
\caption{Final Parameter Settings Used in the Study}
\label{tab:parameters}
\begin{tabular}{lcc}
\textbf{Parameter} & \textbf{Symbol} & \textbf{Value} \\
Smooth Window & $N$ & 3 \\
Engagement Confirmation Window & $W_e$ & 1s \\
Disengagement Confirmation Window & $W_d$ & 1s \\
Engagement Threshold & $\tau_e$ & 40\% \\
Disengagement Threshold & $\tau_d$ & 50\% \\
\end{tabular}
\end{table}

\section{Experiment}
\begin{figure*}
  \centering
  \includegraphics[width=0.92\textwidth]{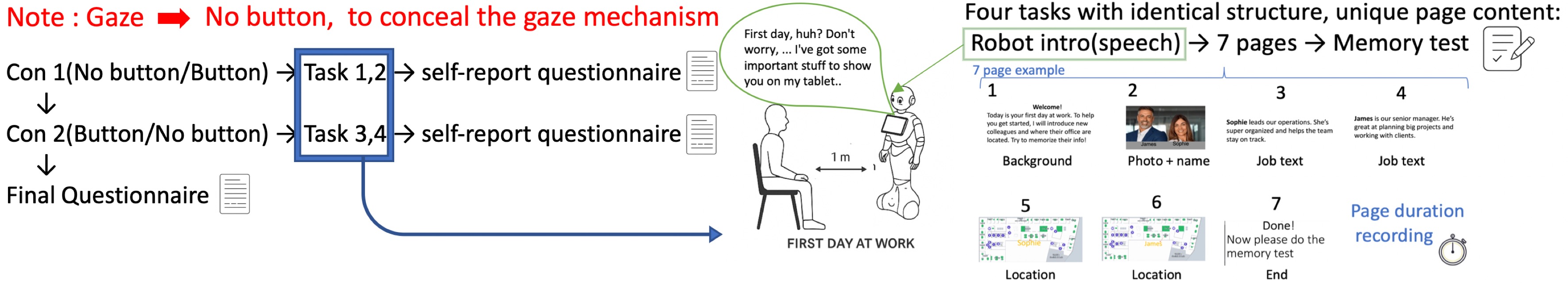}
  \caption{Flowchart of the experimental procedure for two interaction conditions, gaze-based and button-based. The gaze-based condition was presented to participants as a “no-button” interaction to conceal the underlying gaze detection mechanism.}
  \label{fig:experiment}
\end{figure*}

To validate the proposed framework, we designed a scenario titled “First Day at Work”, where a humanoid robot acts as a receptionist welcoming a new employee. During onboarding, the robot introduces two colleagues using its tablet display. We conducted a within-participant study in which participants experienced both conditions while viewing seven information pages (see Figure \ref{fig:experiment}). In the button-based condition, participants tapped a tablet button to advance pages. In the gaze-based condition, our gaze-tracking framework automatically advanced pages once participants completed the visual task. Participants were not informed that their gaze controlled page progression, allowing us to examine whether gaze can support intuitive interaction.


\begin{figure*}
  \centering
  \includegraphics[width=0.9\textwidth]{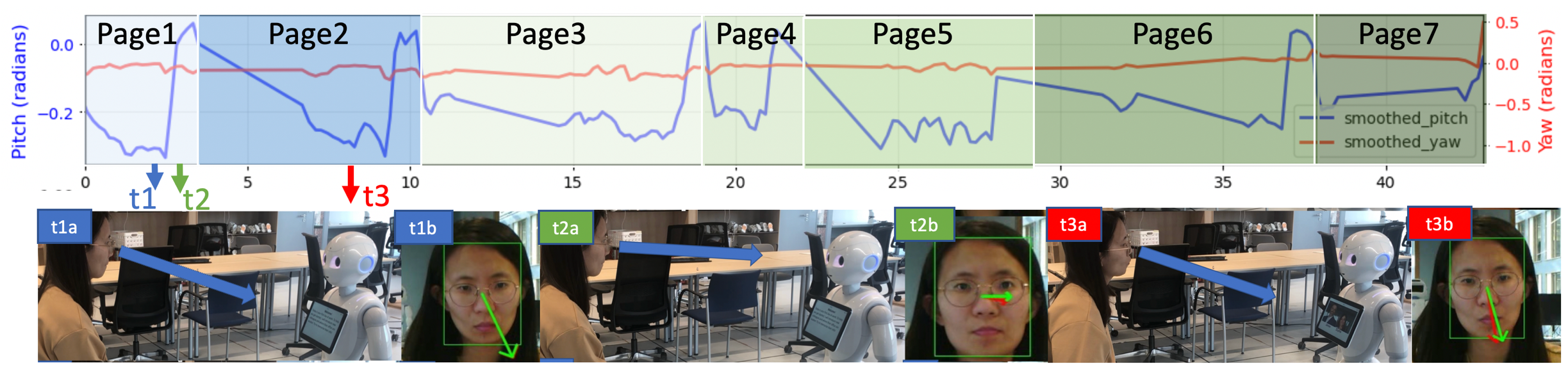}
  \caption{Example of a gaze-based interaction task. \textbf{Top:} Smoothed pitch (blue) and yaw (red) over time. The semi-transparent blue areas indicate periods when the robot displayed Page~1 and Page~2. \textbf{Bottom:} Frame sequences from real-time recordings corresponding to three moments ($t_1$, $t_2$, $t_3$) in the top plot. Panels (a) and (b) show views from a third-person perspective (a) and the robot's perspective (b). In (b), the arrows represent the original 3D gaze (red) and the smoothed gaze (green). These three moments illustrate the transition from Page1 to Page2 using the proposed framework.}
  \label{fig:example}
\end{figure*}

\subsection{Experimental Setup}
The experiment was conducted in a laboratory setting, including a Pepper robot (version 1.8a) \cite{pandey2018mass} equipped with a built-in monocular camera and WiFi connectivity, an OMEN laptop with an RTX 3070 GPU, and a participant. The robot and participant were positioned approximately 100~cm apart. The built-in monocular camera captured video frames containing the participant’s face at a resolution of 640×480 pixels and a frame rate of 10~Hz. These frames were transmitted to the laptop via WiFi, where they were processed by the experimental framework. The resulting robot actions were then sent back to Pepper for execution. Although the camera captured frames at 10~Hz, the system updated robot actions at approximately 5~Hz due to the computation time required for processing each frame and generating appropriate responses. For instance, the L2CS gaze estimation model ran at approximately 126~ms per frame on our setup.

\subsection{Procedure}
Upon arriving at the laboratory, participants were briefed on the experiment and provided informed consent. Participants were then seated approximately one meter in front of a Pepper humanoid robot. The robot introduced itself via speech, stating: "Hey there! I am Pepper. Really nice to meet you. First day, huh? Don't worry, I'll help you get the hang of things. I've got some important stuff to show you on my tablet. When you’re done reading each section, I’ll guide you to what’s next" The robot then guided participants through four scenario tasks equally divided between two interaction conditions: gaze-based interaction and button-based interaction. The order of interaction conditions was counterbalanced across participants, with participants aware there were two interaction conditions, one with a button and another without, but unaware that page progression in the gaze-based condition was triggered by their gaze behavior.

All four tasks were conducted within the “First Day at Work” scenario, as shown in Figure \ref{fig:experiment}, where the robot introduced two new colleagues through seven pages displayed on its tablet: background story (Page 1), colleague photos (Page 2), job descriptions (Pages 3-4), office locations (Pages 5-6), and closing message (Page 7). Between pages, the robot provided verbal transitions such as "Here are their names and photos" before Page 2, "Now, here's what the first colleague does" before Page 3, and "This is what the second colleague does" before Page 4. In the gaze-based interaction, the robot’s head-mounted camera captured facial images and automatically advanced the page when the gaze-tracking system detected task completion, with a 10-second timeout serving as a failsafe. This timeout was chosen empirically based on a pilot study with seven participants, each completing a task of six trials. The maximum page durations observed ranged from 5 to 8 seconds, so the timeout was conservatively set to 10 seconds to ensure reliability. In the button-based interaction, participants manually progressed by clicking a tablet button while the robot's camera recorded for offline analysis without influencing progression. Page viewing durations were recorded in both interaction conditions. After each task, participants completed a short memory test to assess their recall of the page content. Upon completing each interaction condition (i.e., after two tasks), they filled out a brief self-report questionnaire, adapted from \cite{bartneck2009measurement}, to evaluate their subjective experience with that interaction. At the end of the session, participants reported their hypotheses about the no-button (gaze-based) interaction and their views on its timing. See Figure \ref{fig:experiment}.  The four memory tests and all questionnaires are provided in the supplementary materials \footref{github}.

An example of a gaze-based interaction task is shown in Figure \ref{fig:example}. This  shows smoothed 3D gaze direction (yaw and pitch) during a complete interaction task with six page turns. Six snapshots from both third-person and robot perspectives illustrate the transition from Page 1 to Page 2, capturing three key phases: engagement with Page 1, disengagement from Page 1, and re-engagement with Page 2.

\subsection{Participant}
Twenty adults (11 males, 9 females), aged 18–35 years, participated in the study. All had normal or corrected-to-normal vision and normal hearing. Each participant received €10 for their time. The study underwent a self-check and was approved by the Ethics Review Committee of the Faculty of Science (approval code: 2024-022), and informed consent was obtained from all participants prior to the experiment. Three participants were excluded due to data collection errors, leaving 17 participants in the final analysis. Given the unknown effect size, a formal power analysis was not conducted. However, based on a comparable study involving a humanoid robot using human gaze in a collaborative building task with 10 participants \cite{palinko2016robot}, a sample size of 17 was deemed appropriate for the current research.

\section{Result}

\subsection{System Accuracy}

\begin{figure}[h]
  \centering
  \includegraphics[width=0.7\linewidth]{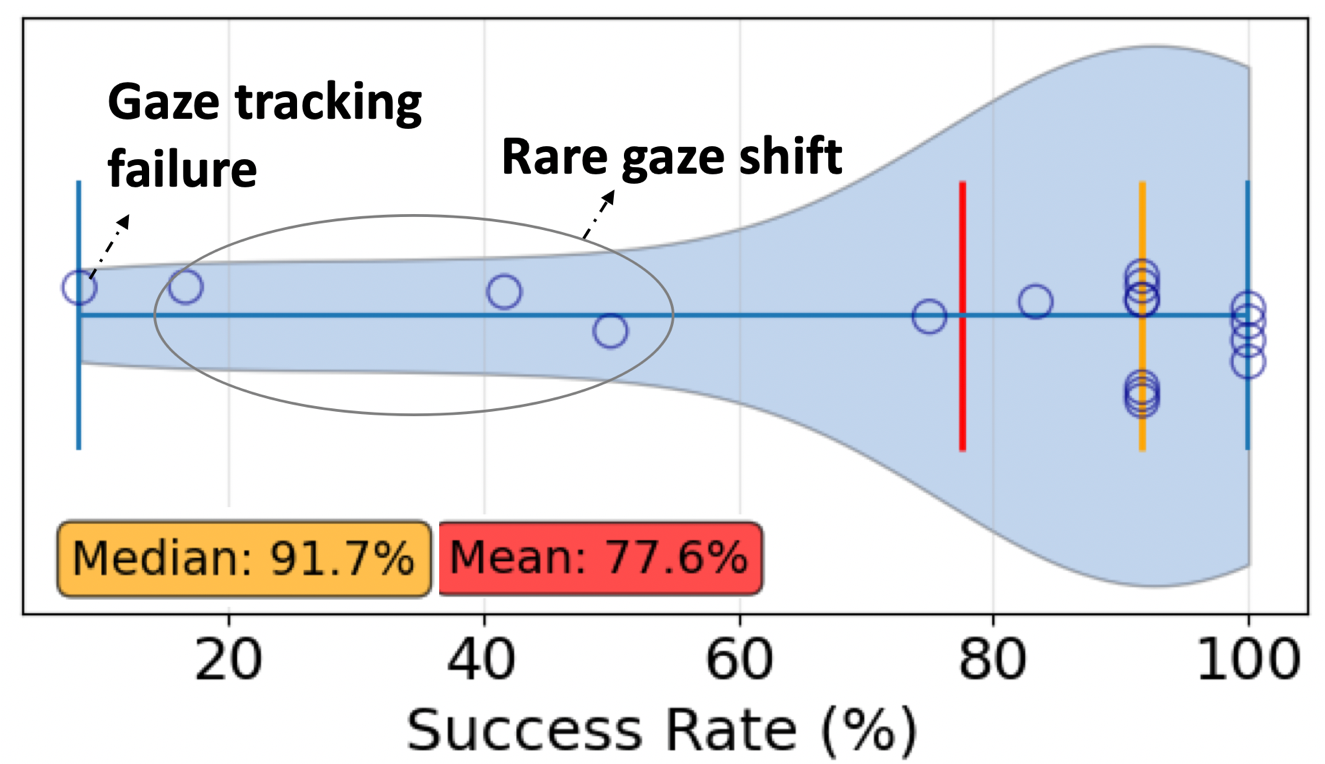}
  \caption{Success Rate Distribution of the Proposed System Across Participants. The violin plot shows the probability density of success rates, with wider sections indicating where more participants cluster. Individual data points (blue dots) represent each participant's success rate, calculated as the percentage of successful state changes detected across all trials. The red line indicates the mean success rate, while the orange line shows the median. The main reasons for the four participants with lower success rates are indicated in the figure with arrows and text.}
  \label{fig:result1}
\end{figure}

The gaze detection system was evaluated with 17 participants, each completing two tasks with six page transitions across seven pages, yielding 204 interaction turns. A turn was considered successful if page advancement was triggered by gaze behavior rather than the 10\,s timeout. Due to minor imprecision in logged times (e.g., 9.8--9.9\,s), success was instead defined by a detected state transition from \textit{engaged} to \textit{disengaged}. As shown in Figure~\ref{fig:result1}, success rates exhibit a bimodal distribution, with a small group achieving low performance (0--50\%) and most participants achieving high performance (80--100\%). The mean success rate was 77.6\% (red line), while the median was 91.7\% (orange line), indicating negative skew caused by a few low-performing participants.

Analysis revealed two main causes for the four participants with success rates below 50\%: (1) \textbf{Gaze tracking failure}, where gaze misclassification occurred due to glasses and sunlight-induced reflections; and (2) \textbf{Rare gaze shift}, where three participants rarely looked back at the robot’s face, with one participant showing almost no gaze transitions, likely because the 10\,s timeout was insufficient for their information processing time, as supported by longer completion times in the button-based condition.

\begin{figure*}
  \centering
  \includegraphics[width=1\textwidth]{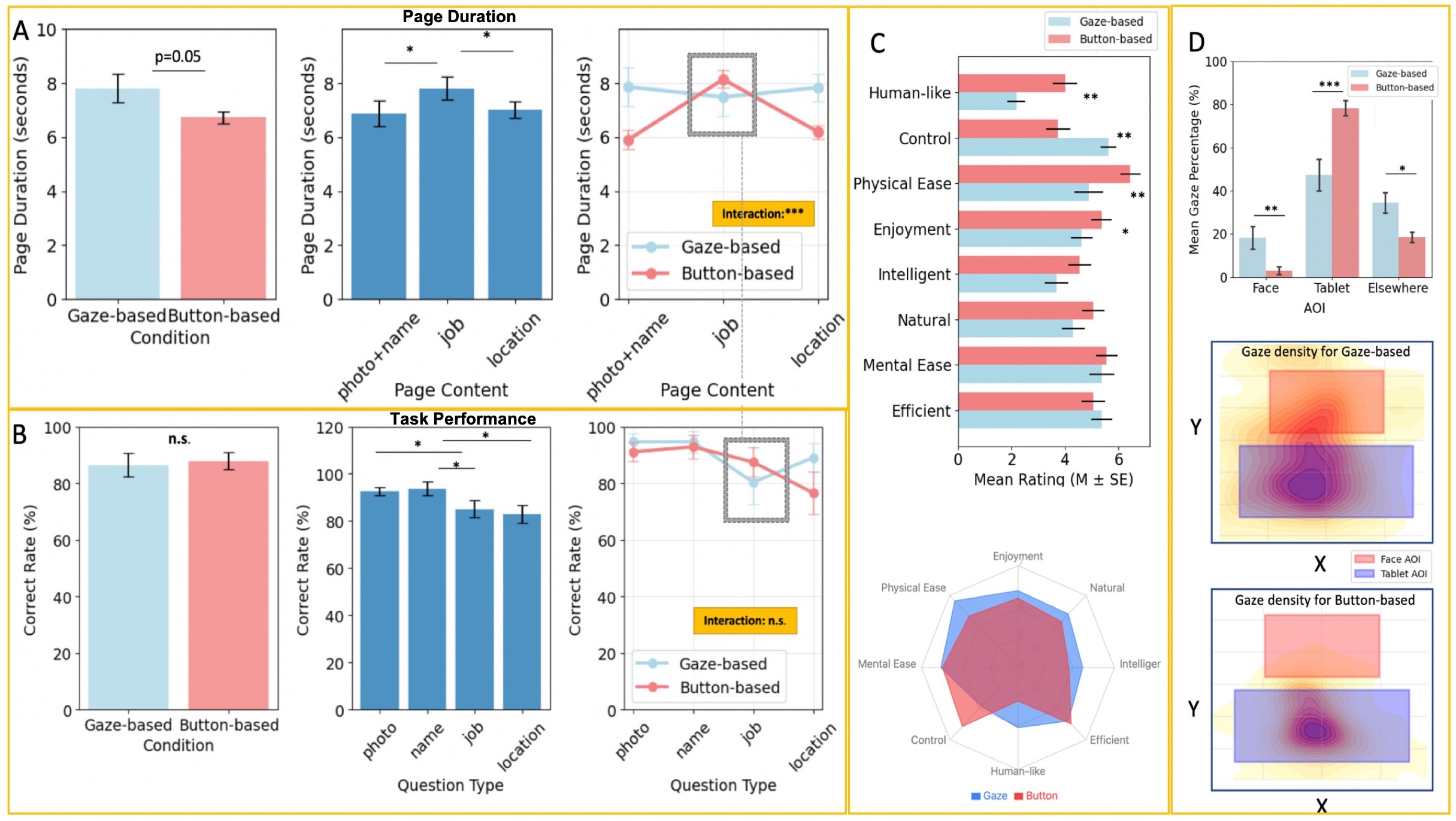}
  \caption{\textbf{Comparison of gaze-based and button-based interaction conditions in memory task.} \textbf{(A) Page duration by condition and content.} Left: Mean duration by input condition. Middle: Mean duration by content type. Right: Condition $\times$ content interaction. \textbf{(B) Memory task accuracy by condition and question type.} Left: Accuracy by input condition. Middle: Accuracy by question type. Right: Condition $\times$ question type interaction. \textbf{(C) User experience.} \textit{Top:} Bar plots showing subjective ratings on a 7-point Likert scale. \textit{Bottom:} Radar chart summarizing mean scores across dimensions. Higher values indicate more positive evaluations. \textbf{(D) Visual attention patterns.} Top: Gaze allocation to face, tablet, and elsewhere regions. Bottom: Spatial gaze density heatmaps by condition.
*p $<$ 0.05, **p $<$ 0.01, ***p $<$ 0.005; n.s., not significant (p $>$ 0.05).}
  \label{fig:comparison}
\end{figure*}

\subsection{Comparison Between Gaze-Based and Button-Based Conditions }

\subsubsection{Page Duration}



Task performance was analyzed for 17 participants completing five question-related pages under two interaction conditions (gaze-based and button-based), yielding 340 trials per condition (17 participants $\times$ 4 tasks $\times$ 5 pages). To account for the 10s timeout in the gaze-based condition versus unlimited time in the button-based condition, button-based trials exceeding 10 s ($n = 27$, 15.9\%) and their corresponding gaze-based trials were excluded. The final dataset comprised 286 trials per condition (84.1\%).

A two-way ANOVA on page duration with factors \textbf{condition} and \textbf{page content} revealed a significant main effect of condition, $F(1,280)=4.7$, $p<.05$, and a significant condition $\times$ page content interaction, $F(2,280)=7.7$, $p<.001$. Post-hoc tests showed marginally longer durations in the gaze-based condition ($M=7.8$ s, $SE=0.5$ s) than in the button-based condition ($M=6.7$ s, $SE=0.2$ s), $t(16)=2.1$, $p=.05$. Job-related pages were significantly longer than both location and photo+name pages (all $p<.05$).

\subsubsection{Memory Task Performance}


After each task, participants answered four question types, yielding 272 responses per condition. Applying the same exclusion criteria, button-based trials exceeding 10 s and their corresponding gaze-based trials were removed, resulting in 242 responses per condition (88.9\%). A two-way ANOVA on accuracy with factors \textbf{condition} and \textbf{question type} showed a significant main effect of question type, $F(3,234)=2.6$, $p<.05$, but no main effect of condition and no interaction. Post-hoc tests revealed higher accuracy for photo and name questions than for job and location questions ($p<.05$), with no differences between interaction conditions.

\subsubsection{Subjective Questionnaire}
Participants rated eight user experience dimensions (7-point Likert scale) for both interaction conditions. Paired t-tests showed that gaze-based interaction was perceived as significantly more human-like ($t(16) = 3.81$, $p < .01$), physically comfortable ($t(16) = 3.13$, $p < .01$), and enjoyable ($t(16) = 2.16$, $p < .05$), but offered less perceived control ($t(16) = -3.58$, $p < .01$). No significant differences were found for Intelligent, Natural, Mental Ease, or Efficient. (See Figure~\ref{fig:comparison}C.)

\subsubsection{Gaze Distribution}

Gaze behavior was analyzed using three AOIs: face, tablet, and elsewhere (Figure~\ref{fig:comparison}D). In the gaze-based condition, 2D gaze points were recorded in real time, whereas in the button-based condition, video recordings were processed offline using the L2CS model to estimate 2D gaze coordinates following the same pipeline. This yielded 6,063 gaze points for gaze-based interaction and 12,630 for button-based interaction, with the difference due to a lower real-time frame rate. As shown in Figure~\ref{fig:comparison}D, participants looked significantly more at the tablet during button-based interaction ($t(16)=4.29$, $p<.001$), and less at the face ($t(16)=-3.11$, $p<.01$) and elsewhere ($t(16)=-2.91$, $p<.05$). Gaze density maps further illustrate these spatial patterns (Figure~\ref{fig:comparison}D, bottom).

\subsection{Natural Gaze Responses in Gaze-Based Interaction}

\begin{figure}[h]
  \centering
  \includegraphics[width=0.8\linewidth]{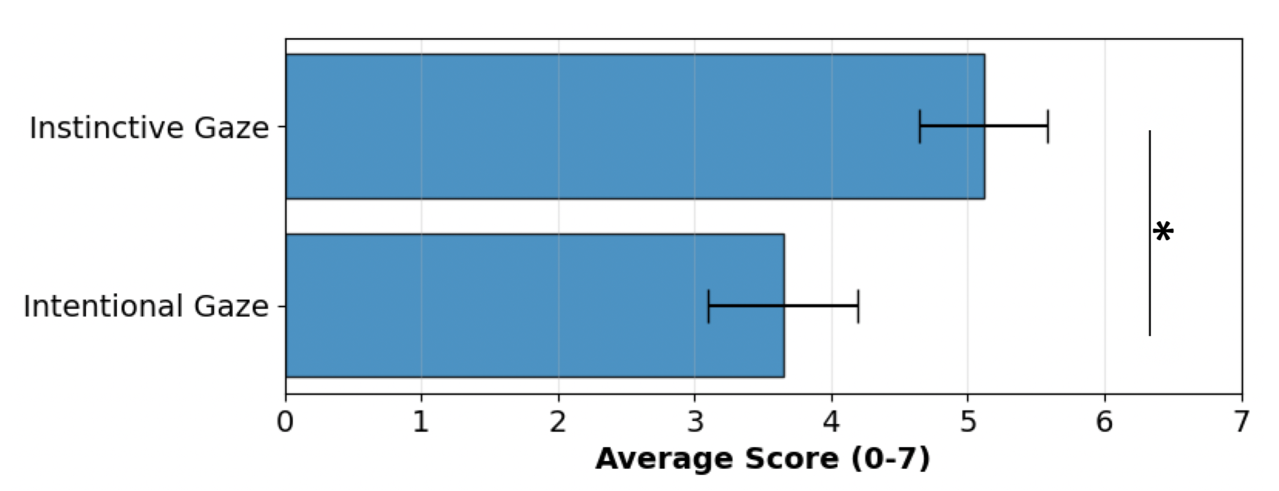}
  \caption{ Self-reported gaze behaviors. * respent $p<0.05$.}
  \label{fig:result3}
\end{figure}

In the final questionnaire, participants rated two statements on a 7-point scale regarding their gaze behavior during the interaction. The results in Figure\ref{fig:result3} show a significant difference between intentional and instinctive gaze behaviors. Participants rated \textit{``I found myself instinctively looking at the robot's face when I finished processing the current page and was ready to continue''} significantly higher ($M = 5.1$) than \textit{``I intentionally looked at the robot's face to trigger a response''} ($M = 3.7$, $p = 0.040$). 


\section{Discussion and Conclusion}

Findings from the \textit{First Day at Work} demonstrate that the presented framework provides a practical, socially grounded approach to gaze-based interaction in HRI:

\begin{enumerate}
    \item \textbf{Practical viability} -- The presented framework achieved a 77.6\% mean success rate, showing that reliable real-time interaction can be achieved without precise calibration. 
    \item \textbf{Behavioral and performance outcomes} -- Participants in the gaze-based condition took slightly longer to complete interactions (\(\approx 1\)~s on average) compared to the button-based condition but showed no difference in task memory performance.
    \item \textbf{User experience} -- Despite reduced control over interaction flow, participants rated the gaze-based interaction as more human-like and physically comfortable. Importantly, instinctive gaze shifts were preferred over intentional signaling, suggesting that gaze-based interaction feels more like an extension of natural social behavior than a learned input skill.
\end{enumerate}

Beyond performance, gaze-based interaction reshaped attention patterns. Participants shifted from a tablet-centered focus 
towards a more distributed attention pattern, encompassing the robot's face.
This shift reframed the experience from device use to social engagement, enhancing perceptions of human-likeness. 
These findings suggest that gaze-based frameworks can support seamless, human-like interaction that is both intuitive and socially grounded.

Despite its promise, several limitations emerged:

\begin{itemize}
    \item \textbf{Layout-related challenges}: Interaction effects emerged between gaze detection and content layout. Job-related text, displayed near the top of the tablet, was sometimes misclassified as face gaze, causing premature page advances and slightly reduced memory accuracy. This indicates that gaze-based readiness detection is most effective when task interfaces are spatially separated from the robot’s face. Future designs should account for spatial configuration to prevent inadvertent gaze misclassification.

    \item \textbf{Technical constraints}: Glasses reflections and strong lighting occasionally caused gaze misclassification. The 10 s failsafe triggered prematurely for slower readers, and some participants did not signal readiness via gaze, challenging the assumption that gaze redirection universally indicates task completion. Adaptive timing and individualized gaze thresholds could mitigate these issues.

    \item \textbf{System limitations}: 
    Parameters were tuned on a small sample, limiting generalizability. Future systems could use data-driven calibration (e.g., Bayesian optimization) and adaptive thresholding. Data loss due to low frame rates, unstable WiFi, and processing delays affected post-hoc analysis, though real-time performance was unaffected. Multi-threaded or distributed architectures could improve reliability. Finally, the system currently operates within approximately 1 meter; integrating depth sensing could extend range and robustness.


\end{itemize}

The presented framework highlights the practical value of off-the-shelf gaze estimation models for enabling intuitive, calibration-free HRI, eliminating the need for specialized hardware. 
By leveraging natural gaze shifts from task interfaces to the robot’s face, the framework formalizes gaze-based turn-taking, using attention transitions as non-verbal cues to signal when control should be returned from the user to the robot. The framework defines a set of parameters that structure this turn-taking process based on gaze estimated from an appearance-based model using face images captured by the robot’s built-in RGB camera. This allows autonomous detection of task completion while preserving task performance and enhancing social engagement. Although slightly slower than button-based input, gaze-based interaction increases perceived human-likeness and intuitiveness, demonstrating that low-cost, RGB-only approaches can support real-time, socially aware HRI.

\addtolength{\textheight}{-12cm}   






\bibliographystyle{IEEEtran}
\bibliography{main}

\end{document}